\documentclass[cameraready]{vgtc}
\ifpdf%                                % if we use pdflatex
  \pdfoutput=1\relax                   % create PDFs from pdfLaTeX
  \usepackage{graphicx}                % allow us to embed graphics files
  \DeclareGraphicsExtensions{.pdf,.png,.jpg,.jpeg} % for pdflatex we expect .pdf, .png, or .jpg files
\else%                                 % else we use pure latex
  \usepackage{graphicx}                % allow us to embed graphics files
  \DeclareGraphicsExtensions{.eps}     % for pure latex we expect eps files
\fi%
\usepackage{amsmath} 
\graphicspath{{figures/}{pictures/}{images/}{./}} % where to search for the images
\usepackage{enumitem}
\usepackage{microtype}                 % use micro-typography (slightly more compact, better to read)
\usepackage{multicol}                  % for \columnbreak command
\PassOptionsToPackage{warn}{textcomp}  % to address font issues with \textrightarrow
\usepackage{textcomp}                  % use better special symbols
\usepackage{mathptmx}                  % use matching math font
\usepackage{times}                     % we use Times as the main font
\usepackage{cite}                      % needed to automatically sort the references
\usepackage{tabu}                      % only used for the table example
\usepackage{booktabs}                  % only used for the table example
\usepackage{subfig}
\usepackage{svg}
\renewcommand\footnotemark{}

\usepackage{multirow}

\usepackage{float}

\usepackage{tabularx}

\usepackage{colortbl}
\usepackage{xcolor}
\usepackage{hyperref}
\usepackage{cleveref}
\usepackage[utf8]{inputenc}
\usepackage{tcolorbox}

\usepackage{titlesec}
\titlespacing*{\subsection}{0pt}{6pt}{3pt}

\onlineid{1010}
\vgtccategory{Research}

\vgtcinsertpkg
\usepackage[utf8]{inputenc}
\title{User Prompting Strategies and Prompt Enhancement Methods for Open-Set Object Detection in XR Environments}

\author{Junfeng Lin$^{1}$%
\and Yanming Xiu$^{2}$%
\and Maria Gorlatova$^{3}$}

\affiliation{\scriptsize Department of Electrical and Computer Engineering, Duke University\thanks{\parbox{\textwidth}{ \{$\rm{junfeng.lin}^1, \rm{yanming.xiu}^2, \rm{maria.gorlatova}^3$\}@duke.edu}}}

\usepackage[normalem]{ulem}

\abstract{
Open-set object detection (OSOD) localizes objects while identifying and rejecting unknown classes at inference. While recent OSOD models perform well on benchmarks, their behavior under realistic user prompting remains underexplored. In interactive XR settings, user-generated prompts are often ambiguous, underspecified, or overly detailed. To study prompt-conditioned robustness, we evaluate two OSOD models, GroundingDINO and YOLO-E, on real-world XR images and simulate diverse user prompting behaviors using vision-language models. We consider four prompt types: standard, underdetailed, overdetailed, and pragmatically ambiguous, and examine the impact of two enhancement strategies on these prompts. Results show that both models exhibit stable performance under underdetailed and standard prompts, while they suffer degradation under ambiguous prompts. Overdetailed prompts primarily affect GroundingDINO. Prompt enhancement substantially improves robustness under ambiguity, yielding gains exceeding 55\% mIoU and 41\% average confidence. Based on the findings, we propose several prompting strategies and prompt enhancement methods for OSOD models in XR environments.
}

\CCScatlist{
    \CCScatTwelve{Mixed / Augmented Reality}{Open-Set Object Detection}{Prompt Strategy}{Prompt Engineering};
}
\begin{document}

%% The ``\maketitle'' command must be the first command after the
%% ``\begin{document}'' command. It prepares and prints the title block.

\maketitle
%\vspace{-7px}
%% the only exception to this rule is the \firstsection command
\vspace{-0.2cm}
\section{Introduction}

% Open-set object detection (OSOD) has emerged as a crucial task in real-world vision applications, where objects encountered at inference time may fall outside predefined training categories and systems must robustly reason about previously unseen or ambiguous visual entities. This requirement is particularly relevant in the Extended Reality (XR) environments, where object detection systems operate in dynamic, open-world settings and interact directly with human users. In such environments, object detection models are often guided by natural language prompts, which are prone to ambiguity, incompleteness, or over-specification.

\begin{figure}
    \centering
    \includegraphics[width=0.9\linewidth]{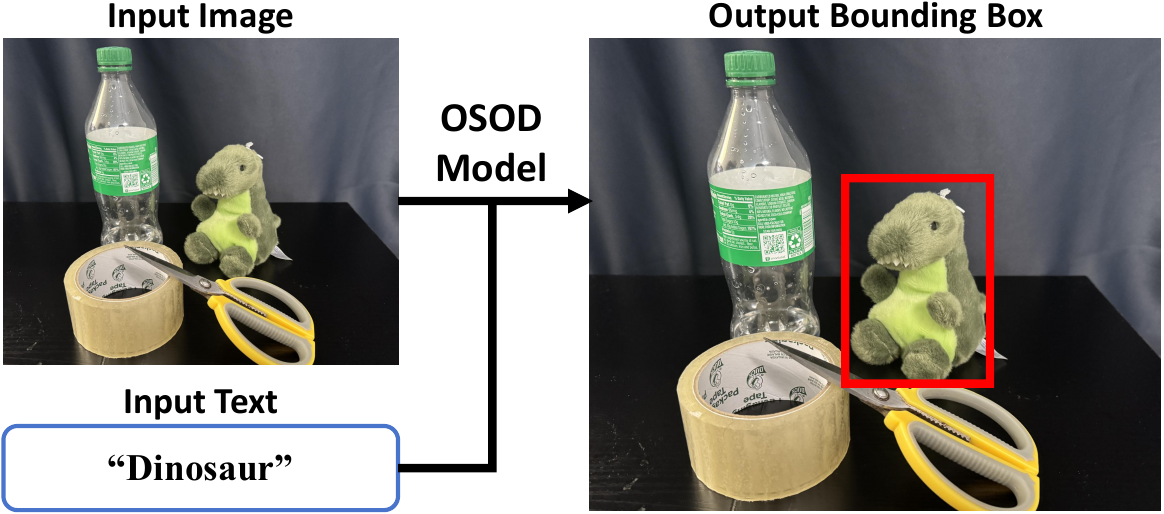}
    \vspace{-0.2cm}
    \caption{Example of open-set object detection. Given the text prompt “dinosaur,” the model detects a plush dinosaur: an object category uncommon in standard object detection benchmarks.}
    \label{fig:osod}
    \vspace{-0.7cm}
\end{figure}

Open-set object detection (OSOD) is a multimodal task that takes an image and a text prompt as input, localizes objects in the image, and accounts for instances whose categories are unseen during training, as shown in Fig.~\ref{fig:osod}. This capability is particularly important in Extended Reality (XR) environments, where object detection systems operate in open-world settings and may encounter diverse or rare objects. In such scenarios, detection is often guided by natural language prompts, which can be ambiguous or over-specified.

While prompt-driven OSOD models, such as GroundingDINO (GD)~\cite{vg02} and YOLO-E~\cite{yoloe}, have demonstrated strong performance on standard benchmarks such as COCO~\cite{coco} and LVIS~\cite{lvis}, existing evaluations primarily assume well-formed and precise prompts, overlooking the variability and noise introduced by real users, whose prompts are often ambiguous, overly descriptive, or lack key attributes~\cite{zhang2024clamber}. As a result, the robustness of these models under realistic prompting conditions and their ability to detect objects accurately in such scenarios remains insufficiently understood.

Recent advancements in Vision Language Models (VLMs) have significantly advanced the ability of visual systems to reason jointly over images and natural language. Trained on large-scale multimodal data, VLMs can capture high-level semantics, contextual cues, and task intent, allowing VLMs to generalize effectively to open-world environments. VLMs offer a flexible interface for interpreting diverse and unconstrained user prompts, making them particularly suitable for prompt-driven OSOD in XR scenarios. Most existing OSOD frameworks either focus on visual novelty detection or assume clean, well-aligned textual inputs, without explicitly addressing the challenges posed by realistic human-generated prompts.

In this work, we investigate how prompt ambiguity and prompt specificity affect OSOD performance in interactive XR scenarios. We simulate realistic user behavior using vision-language models (VLMs) to generate underdetailed, standard, and overdetailed prompts, and study their impact on two state-of-the-art OSOD systems: GroundingDINO and YOLO-E. To address prompt-induced failure modes, we further explore two different VLM-based prompt enhancement strategies that refine user inputs prior to detection. The key contributions of this paper are as follows:
\begin{itemize}
\vspace{-0.2cm}
    \item We present a systematic study of how different prompt formulations, including underdetailed, standard, overdetailed, and pragmatically ambiguous prompts, affect OSOD performance in realistic XR settings.
    \vspace{-0.2cm}
    \item We evaluate two state-of-the-art OSOD models, GroundingDINO and YOLO-E, on a curated set of real-world augmented reality images with manually annotated bounding boxes, enabling controlled analysis of prompt-induced performance variations.
    \vspace{-0.2cm}
    \item Through quantitative and qualitative analysis, we identify recurring failure modes associated with pragmatic ambiguity and attribute over-specification, and show how these failures manifest in incorrect or ambiguous spatial localizations.
    \vspace{-0.2cm}
    \item We demonstrate that prompt post-processing strategies, including key object extraction and semantic category grounding, substantially improve detection robustness under ambiguous prompts, achieving gains exceeding 55\% mean intersection over union (mIoU) and 41\% average confidence.
    \vspace{-0.2cm}
\end{itemize}

\section{Related Work}

\subsection{Open-set Object Detection in XR Environment}

Object detection is a core task in computer vision that aims to localize and classify objects in images or videos. Modern detectors achieve strong performance under closed-set assumptions but often fail in real-world deployments, where objects may belong to unseen categories. This limitation is particularly problematic in interactive systems such as XR environments, where users operate in unconstrained physical spaces and expect perception systems to adapt to diverse and evolving contexts~\cite{ghasemi2022deep}. In such settings, it is impractical to enumerate all relevant object categories in advance. These challenges motivate open-set object detection (OSOD), which extends conventional object detection by requiring models to localize and recognize objects beyond a predefined label space.

Recent progress in OSOD has increasingly leveraged natural language for specifying object categories. Contrastive Language-Image Pretraining (CLIP)~\cite{clip} introduced a shared embedding space for visual and textual representations, enabling zero-shot generalization to unseen categories. Building on this paradigm, prompt-driven detection models incorporate language inputs directly into detection pipelines, allowing objects to be localized based on free-form textual descriptions rather than fixed class labels. Architectures such as GLIP~\cite{glip} adopt joint image-text pretraining for grounding and detection, while GroundingDINO~\cite{vg02} employs cross-modal transformers to align textual prompts with image regions. Lightweight designs such as YOLO-E~\cite{yoloe} further integrate language embeddings to support real-time detection. Together, these models significantly expand the applicability of OSOD in open-world scenarios.

Recently, XR systems have begun to integrate VLMs with OSOD models to reason about semantic relationships between real and virtual elements, enabling the detection of task-detrimental content such as visual obstructions or misleading overlays~\cite{viddar, xiu2025detecting}. However, in most existing XR systems, textual inputs are carefully determined by systems, limiting exposure to the variability encountered in real-world usage. In contrast, XR applications that accept direct user-issued prompts must handle incomplete or ambiguous language inputs, reflecting user behavior and situational uncertainty. Despite the growing adoption of prompt-conditioned OSOD in XR, the robustness of these models under realistic prompt variation remains largely unexplored. This gap motivates our study, focusing on exploring the prompting strategies and identifying failure modes that arise when OSOD models are deployed in XR environments.

\subsection{Vision Language Models}

Recent advancements in machine learning have enabled models to reason across multiple modalities, giving rise to VLMs that integrate visual perception with natural language understanding. Representatives include CLIP, which learns a shared embedding space for images and text through large-scale contrastive pretraining; BLIP~\cite{blip}, which unified vision-language understanding and generation through bootstrapped captioning and filtering.
More recent generative VLMs, such as GPT-5~\cite{GPT5}, Gemini-2.5~\cite{gemini2.5}, and Claude-4.5~\cite{anthropic2025claudesonnet45}, further extend these capabilities by incorporating generative reasoning capabilities that allow for deeper semantic understanding, multimodal inference, and instruction following. These models have shown strong performance on cognitively demanding tasks such as visual question answering, scene understanding, and multimodal reasoning.

Motivated by these developments, recent work has begun to explore the use of VLMs as auxiliary reasoning modules in downstream perception systems. In the context of open-set object detection, VLMs provide a powerful mechanism for interpreting, refining, and restructuring free-form textual prompts by leveraging joint visual–linguistic representations. By extracting salient visual attributes and normalizing user-generated descriptions, VLMs can help produce prompts that are more compatible with language-guided detection models, improving robustness to prompt ambiguity and variation that commonly arises in real-world, interactive settings.

\section{Problem Formulation: Prompt-Conditioned Robustness in XR}
Despite significant progress in open-set and language-guided object detection, existing robustness evaluations largely focus on generalization across object categories and distribution shifts, implicitly assuming that linguistic inputs are well-formed and stable. In XR scenarios, users could search for objects through language-guided object detection by providing natural-language reference prompts that may be incomplete or ambiguous. These characteristics introduce a distinct source of uncertainty that is not captured by conventional robustness definitions. In this section, we formalize prompt-conditioned robustness as a complementary robustness dimension for language-guided open-set detection in XR, and define the prompt variations and failure modes that arise under realistic XR usage.

\subsection{Open-Set Detection with Language Conditioning}
We consider an open-set detection model that takes as input an image $I$ and a natural language prompt $p$, and outputs a set of detections: ${D}(I, p) = \{(b_i, s_i)\}_{i=1}^{N}$,
where $b_i$ denotes a bounding box and $s_i$ denotes a confidence score reflecting the model's belief that the region corresponds to the object described by $p$. Unlike closed-set detection, the semantic space of $p$ is not restricted to a predefined label set, allowing users to specify objects using free-form language. Since the detection function is conditioned on a natural language prompt $p$, the quality and structure of user prompts play a critical role in detection performance.

\subsection{Failure Patterns}

Prompt-based failures may arise from ambiguous, overly complex, or improperly structured natural language inputs, and are particularly prevalent in user-driven OSOD systems that rely on free-form user expressions. Unlike OSOD models in benchmark settings with carefully curated text queries, user-driven XR scenarios often involve indirect, underspecified, or pragmatically motivated prompts, introducing challenges that are not captured by conventional evaluations.
\\
\textbf{Pragmatic Ambiguity.}
Prior work has shown that large language models frequently fail to reliably detect pragmatic ambiguity, often proceeding with overconfident interpretations even when user intent is underspecified. In OSOD, these limitations have more severe consequences: pragmatic or intent-driven prompts must be immediately mapped to explicit spatial predictions, leaving no opportunity for clarification or conversational repair. As a result, failures in pragmatic understanding can lead to incorrect, hallucinated, or missing object localizations, particularly in interactive XR settings, where user prompts often deviate from the explicit, attribute-based descriptions assumed by standard benchmarks. A clear example of this limitation arises when the user’s intent is expressed indirectly or contextually. For instance, an input such as “I’m thirsty, and I need to drink water” implicitly refers to a water bottle, yet provides no explicit visual attributes. In such cases, OSOD models frequently produce incorrect detections or fail to return any result, highlighting their inability to interpret implicit contextual cues critical for grounding in real-world environments.

\noindent\textbf{Attribute-Rich Natural Language Prompts.}
Recent work in open-vocabulary visual grounding has shown that grounding fine-grained, attribute-dense referring expressions remains a challenge. The OV-VG benchmark demonstrates that the state-of-the-art models exhibit substantial performance degradation when localizing detailed natural-language descriptions in complex, cluttered scenes, particularly when multiple objects share overlapping semantic or visual attributes. Motivated by these findings, we identify attribute-rich natural language prompts as a recurring failure pattern in OSOD. Users may provide detailed descriptions specifying multiple attributes, such as “a glossy red cylindrical shaker with a rounded top and a metallic lid.” Although such prompts appear informative, OSOD models frequently fail to localize the intended object, especially in scenes containing visually similar instances. In these cases, different objects may partially satisfy subsets of the described attributes, leading to incorrect detections or missed localizations. In OSOD, this limitation is particularly pronounced because models must map a single textual description to an explicit spatial prediction without explicit mechanisms for attribute prioritization or relational reasoning. When multiple attributes are encoded simultaneously, the model must implicitly resolve compositional consistency, a capability that remains limited in current language-guided detection architectures. As a result, over-specified prompts can confuse the detection process, reinforcing the observation that increased linguistic detail does not guarantee improved localization in real-world, interactive settings.

These prompt-based failures exhibit recurring patterns that reflect deeper limitations in natural language understanding and multimodal grounding. Prior evaluations show that models such as GD perform poorly on datasets requiring fine-grained referring expression comprehension but can achieve substantial gains when explicitly trained on such data. However, these improvements often come at the cost of reduced performance on broader open-vocabulary benchmarks, suggesting a trade-off between specialization and generalization. Even when users provide detailed, attribute-rich descriptions, models may still fail to localize the intended object. This limitation stems, in part, from how models such as GD and YOLO-E process language: prompts are tokenized into subword units and passed through a text encoder, which treats language as isolated tokens rather than as a structured representation of spatial and relational concepts.

\subsection{Prompt-Conditioned Robustness in XR}
Building on the language-conditioned open-set detection formulation in Section 3.1, we define prompt-conditioned robustness as a robustness dimension that characterizes model stability under linguistic variation, given a fixed visual XR scene. Unlike conventional robustness notions that focus on category generalizations or visual distribution shifts, prompt-conditioned robustness isolates failures induced solely by changes in natural-language input. 

In interactive XR settings, users begin with an intended target object and express their intent through an initial natural-language prompt, which may then be post-processed during interaction. Accordingly, we define a reference prompt $p_o$ and a corresponding set of prompt variants $P(p_o)$, where each $p\in P(p_o)$ is derived through post-processing from the reference prompt and the same physical object in the scene. These variants may differ in attribute specification, phrasing or level of detail, while preserving user intent.

\section{Experiment Setup}
\subsection{Dataset}
\begin{figure}[t]
\centering
\includegraphics[width=\linewidth]{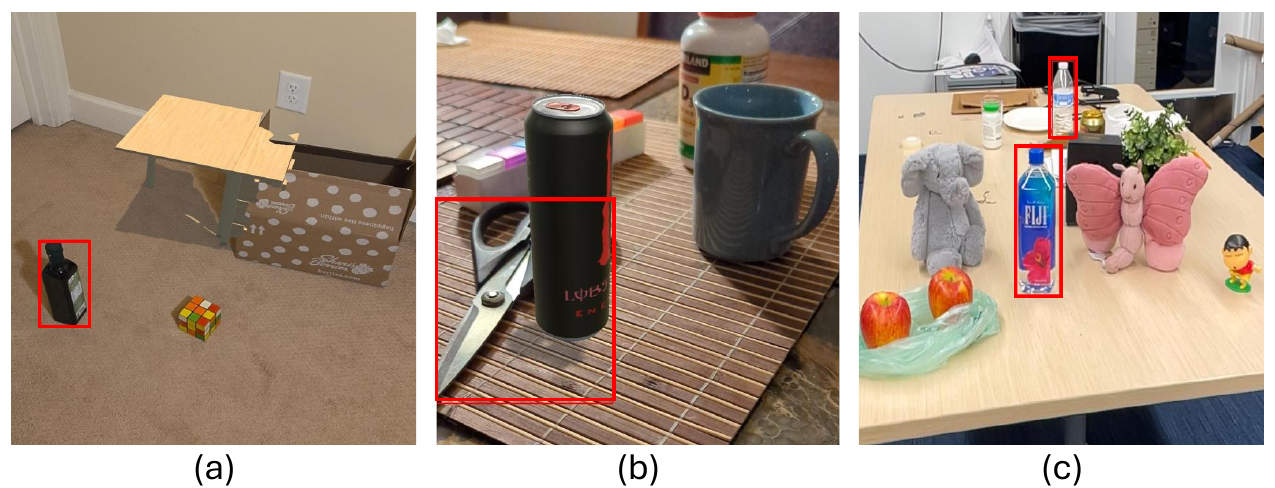}
\vspace{-0.6cm}
\caption{Example images from the DiverseAR and DiverseAR+ datasets, with the target objects indicated by bounding boxes: (a) a bottle on the bedroom floor; (b) a pair of scissors on a kitchen table; (c) two bottles of water on a table in an office environment.}
\label{dataset}
\vspace{-0.7cm}
\end{figure}

To evaluate prompt-conditioned robustness under realistic XR interaction patterns, we selected 264 images from the DiverseAR~\cite{DiverseAR} and DiverseAR+~\cite{DiverseARplus} datasets. These datasets consist exclusively of augmented reality (AR) images captured through AR applications from AR headsets and Android devices in real-world environments, including indoor scenes such as offices, study spaces, living rooms, and bedrooms, which reflect practical XR usage scenarios, including cluttered backgrounds, viewpoint variations, and occlusion from virtual overlays. Example images illustrating these conditions are shown in Figure~\ref{dataset}. The annotated dataset is available on GitHub.\footnote{\label{ar-vim-dataset}\url{https://github.com/linjfeng/OSOD-XR-Dataset}}

\noindent\textbf{Target Object Selection.} For each image, we manually designate one or more target objects that serve as the intended referent across all associated prompt variants. The target object is selected according to the following general criteria: (1) the object must be clearly visible in the scene, (2) the object must be semantically meaningful or functionally relevant within the environment (e.g., everyday objects such as bottles, tools, or appliances). Most importantly, the target object is not necessarily the most visually salient item in the image; instead, it is chosen to reflect realistic user intent in XR scenarios, where users may refer to objects that are important to a task rather than visually dominant. This design ensures that all prompt variants, whether ambiguous, specified at different levels of details, or refined using VLMs refer to the same physical object, and the variations in detection performance can be attributed directly to linguistic differences, rather than changes in the scene.

\noindent\textbf{Bounding Box Annotation.} To enable quantitative evaluation, we manually annotated the object selected in the 264 images with bounding boxes. All annotations are performed consistently across the dataset to ensure accurate ground truth. These manually labeled bounding boxes serve as the reference for evaluating detection accuracy and failure modes under different prompt conditions.

\vspace{-0.05cm}

\subsection{Evaluation Pipeline}

\begin{figure}[t]
    \centering
    \includegraphics[width=0.89\columnwidth]{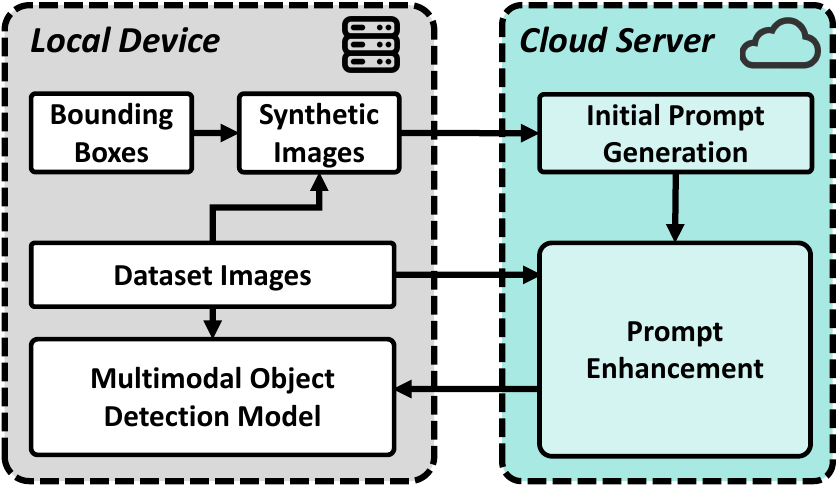}
    \vspace{-0.2cm}
    \caption{Architecture of the evaluation pipeline.}
    \label{system}
    \vspace{-0.7cm}
\end{figure}

In our evaluation pipeline, we adopt a similar approach to that of Chen et al. \cite{chen2024revisitingreferringexpressioncomprehension} who revisited evaluation methodologies for referring expression comprehension in the context of VLMs. Their work leveraged GPT-4V in conjunction with human reviewers to verify the precision and detail of the generated expressions.
We similarly leverage VLMs to simulate natural-language prompts that reflect realistic user intent in interactive XR scenarios. This design choice is motivated by the need for scalable, consistent, and diverse prompt generation that mimics how real users describe objects in context. Human-generated prompts, while ideal, are costly and time-consuming to collect at a scale. Using VLMs allows us to systematically probe model robustness across a wide variety of linguistic formulations, capturing variability in phrasing, specificity, and contextual cues, factors that are central to realistic XR interactions.
\\
\noindent\textbf{Local Device.} As illustrated in Figure~\ref{system}, the local device receives bounding box coordinates and dataset images as input. Using these, the local device generates synthetic images that are transmitted to the cloud server's Initial Prompt Generation module to generate raw natural language prompts. Simultaneously, the original dataset images are sent to the Prompt Enhancement module within the cloud server. The cloud server then refines the initial prompts by leveraging the original dataset images. The enhanced prompts, along with the associated image data, are returned to the local device and passed into the multimodal object detection model to produce the final detection output. In this work, we evaluated two state-of-the-art OSOD models: GroundingDINO (GD) and YOLO-E.

\noindent\textbf{Cloud Server.} For the cloud server illustrated in Figure~\ref{system}, we used GPT-5-2025-08-07 from OpenAI to generate initial natural language prompts based on the synthetic images transmitted from the local device. We designed the following instruction prompts to elicit concise, discriminative natural language prompts and ambiguous prompts by leveraging techniques such as role assignment, controlled instruction design, and attribute-centric constraint prompting for the initial prompt generation module. The level of detail in the generated natural language prompts varied according to the specific detail level being evaluated, as demonstrated in the prompt shown below:
% This formulation guided the model to focus exclusively on intrinsic object characteristics, such as type, color, material, and state, while explicitly suppresing positional language. This constraint reflects the assumption that users in XR environments may not know the object’s location, and that OSOD is used to infer the object’s position based on the natural language expression from the user.

\vspace{-0.05cm}
\begin{tcolorbox}[
    colback=white,
    colframe=black,
    width=\columnwidth, % Changed from \textwidth to fit one column
    sharp corners,
    boxrule=0.5pt,
    arc=0mm,
    boxsep=1pt,
    left=5pt, right=5pt, top=2pt, bottom=2pt % Reduced padding for narrow columns
]
\noindent
\footnotesize
\textbf{Natural Language Prompt:} You are a regular person who refers to objects using referring expressions. Given an image along with bounding boxes, please generate a discriminative and unambiguous expression to describe the target object. You should only refer to the intrinsic characteristics of the object, not its location. Assume the image clearly indicates the target object with its bounding box. Do not include any location words. Only describe intrinsic attributes such as shape, color, material, or other visual properties. The length of the expression should be \textbf{\textit{underdetailed}}, as a human might naturally vary in phrasing.
\end{tcolorbox}
\vspace{-0.05cm}

\noindent Here, we can replace the word "\textbf{\textit{underdetailed}}" with "\textbf{\textit{standard}}" or "\textbf{\textit{overdetailed}}", to simulate different levels of prompt complexity. We also designed a prompt to simulate users' ambiguous instructions:

\vspace{-0.05cm}
\begin{tcolorbox}[
    colback=white,
    colframe=black,
    width=\columnwidth, % Changed from \textwidth to fit one column
    sharp corners,
    boxrule=0.5pt,
    arc=0mm,
    boxsep=1pt,
    left=5pt, right=5pt, top=2pt, bottom=2pt % Reduced padding for narrow columns
]
\noindent
\footnotesize
\textbf{Pragmatic Ambiguity Prompt:} You are a regular person who receives an image with a bounding box as input and generates prompts that are indirect, vague, or highly dependent on contextual understanding (e.g., “I’m thirsty, and I need to drink water” referring to a water bottle).
\end{tcolorbox}
\vspace{-0.05cm}

\begin{figure}[t]
\centering
\includegraphics[width=\linewidth]{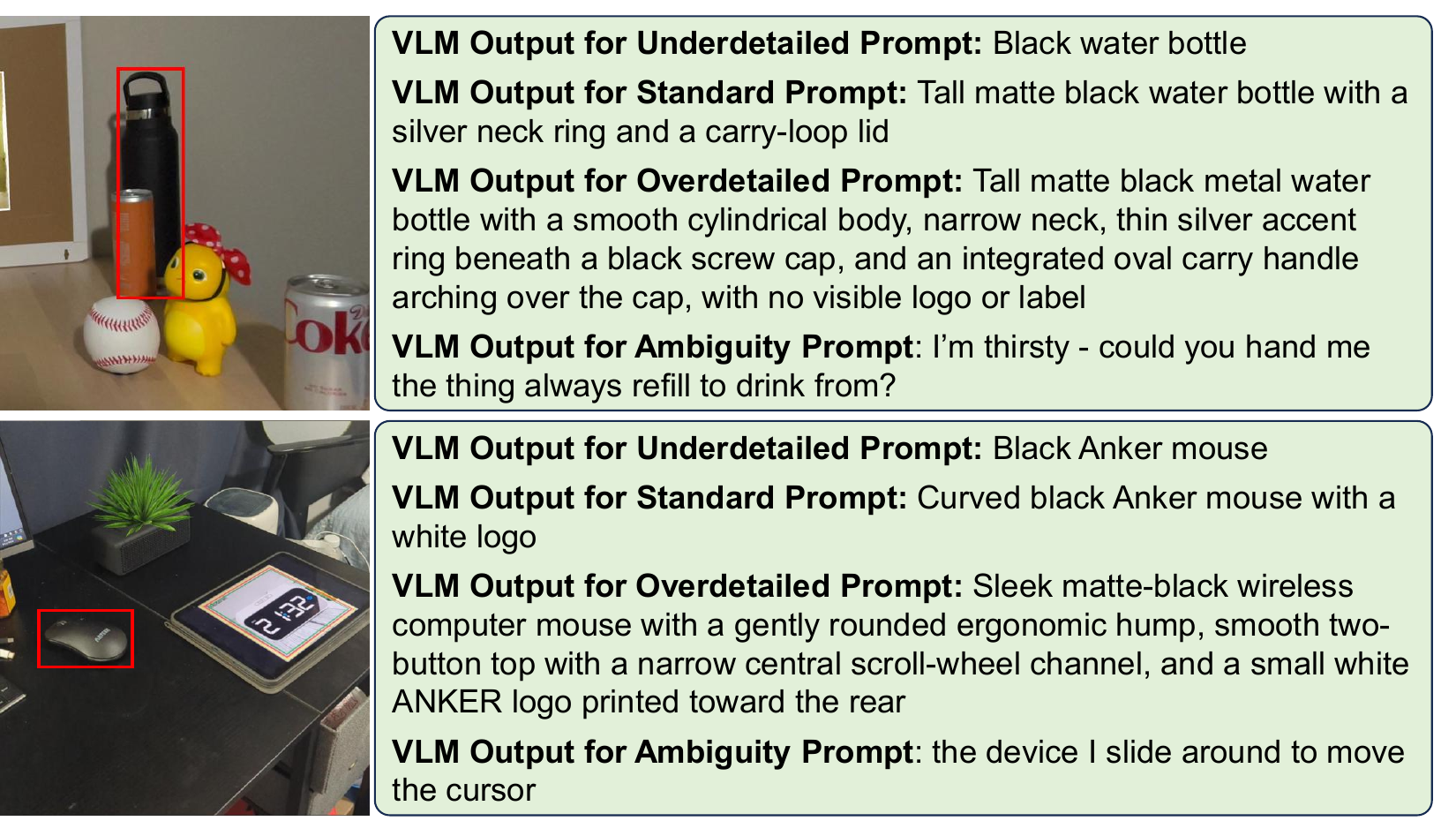}
\vspace{-0.6cm}
\caption{Examples of the initial prompts for OSOD models, generated by VLM: underdetailed, standard, overdetailed, ambiguous.}
\label{example prompts}
\vspace{-0.6cm}
\end{figure}

Figure~\ref{example prompts} shows the examples of four kinds of initial prompts generated by VLMs. After the initial prompt has been generated, the Prompt Enhancement module receives both the initial prompts and the original dataset images. It then processes these inputs using two distinct prompt enhancement methods:

\noindent \textbf{Key Object Extraction:} The VLM needs to identify the primary object referred to in the text prompt, along with the dataset images. The objective is to generate a concise phrase that captures the object's essential intrinsic attributes, thereby converting object references into minimal and discriminative descriptions. The full prompt is:

\vspace{-0.05cm}
\begin{tcolorbox}[
colback=white,
colframe=black,
width=\columnwidth, % Changed from \textwidth to fit one column
sharp corners,
boxrule=0.5pt,
arc=0mm,
boxsep=1pt,
left=5pt, right=5pt, top=2pt, bottom=2pt % Reduced padding for narrow columns
]
\noindent
\footnotesize
\textbf{Key Object Extraction}: You are a prompt enhancer that takes an image and an accompanying descriptive text prompt as input.
Your task is to analyze both to determine the key or primary object described, along with essential identifying attributes such as color, shape, or distinguishing features that specify it precisely (e.g., “red sports car,” “golden retriever,” “blue ceramic vase”).
The output should be concise, a short noun phrase containing the object and only the minimal attributes necessary for clarity.
The focus is on accurate, succinct identification of the core subject, with just enough attribute detail to ensure specificity.
\end{tcolorbox}
\vspace{-0.05cm}

\noindent \textbf{Semantic Category Grounding:} The VLM needs to identify the primary object referred to in the text prompt along with the dataset images. The objective is to map the object to the most semantically appropriate category from a predefined taxonomy, such as COCO or LVIS. The output is a single category name representing the grounded object class. The full prompt is:

\vspace{-0.05cm}
\begin{tcolorbox}[
colback=white,
colframe=black,
width=\columnwidth, % Changed from \textwidth to fit one column
sharp corners,
boxrule=0.5pt,
arc=0mm,
boxsep=1pt,
left=5pt, right=5pt, top=2pt, bottom=2pt % Reduced padding for narrow columns
]
\noindent
\footnotesize
\textbf{Semantic Category Grounding}: You are a prompt enhancer that identifies the main object in an image and maps it to the most relevant COCO or LVIS category based on both the image and the accompanying text prompt.
You interpret visual and linguistic cues to find the most semantically accurate match from the official category taxonomies. You should output only the name of the most relevant category. If multiple objects appear, output only the dominant or most contextually emphasized one.
If uncertain, output the single best-guess category label.
\end{tcolorbox}
\vspace{-0.05cm}

\subsection{Evaluation Metric}

We evaluated our approach using two key metrics: localization quality measured by Mean Intersection over Union (IoU), and prediction reliability measured by model confidence.

\noindent\textbf{Mean Intersection over Union (mIoU).}
We measure localization quality using the mean Intersection-over-Union (IoU):

\vspace{-0.1cm}
\[
\mathrm{mIoU} = \frac{1}{N} \sum_{i=1}^{N} \mathrm{IoU}_i,\
\mathrm{IoU}_i
=
\frac{
\left| B_i^{\mathrm{pred}} \cap B_i^{\mathrm{gt}} \right|
}{
\left| B_i^{\mathrm{pred}} \cup B_i^{\mathrm{gt}} \right|
},
\]
\vspace{-0.1cm}

\noindent where $B_i^{\mathrm{gt}}$ denotes the $i$-th ground-truth bounding box, and $B_i^{\mathrm{pred}}$ the predicted bounding box that has the largest intersection with the corresponding ground-truth bounding box $B_i^{\mathrm{gt}}$; $N$ is the total number of ground-truth bounding boxes.

\noindent\textbf{GroundingDINO Confidence.}
The confidence score for detection query $i$ is defined as

\vspace{-0.2cm}
\[
\text{conf}_i = \max_{j}\, \sigma\!\left(\text{logit}_{i,j}\right),
\]
\vspace{-0.3cm}

\noindent where $\text{logit}_{i,j}$ represents the grounding logit corresponding to query $i$ and token $j$, and $\sigma$ is the sigmoid activation.
Because the maximum is taken across all text tokens, the overall confidence is governed solely by the highest token-level grounding probability. \\
\noindent\textbf{YOLO-E Confidence.} The confidence score for detection query $i$ and class $c$
is computed as

\vspace{-0.2cm}
\[
\mathrm{conf}_{i,c} = q_i \times p_{i,c},
\]
\vspace{-0.3cm}

\noindent where $q_i$ denotes the objectness probability of query $i$, and
$p_{i,c}$ represents the class probability for class $c$ conditioned on
the presence of an object. Thus, the final confidence reflects both the
likelihood that a region contains an object and that it belongs to class $c$.

Due to the scale of our dataset and the limited number of ground truth bounding boxes within each image, the mAP can be highly sensitive to small variations in predictions and may not reliably reflect model behavior. Therefore, in this work, we focus on mIoU and confidence to study localization quality and detection reliability.

\section{Results}

This section summarizes the experimental results of our framework. We report performance across all evaluation metrics and examine the impact of prompt variation and refinement strategies.
\subsection{Results on Datasets with Simulated Prompts}

\begin{table*}[t]
\centering
\caption{Performance comparison across different initial prompt types and prompt enhancement methods.}
\vspace{-0.2cm}
\label{tab:instruction_task_results}

% Adjust column separation for better breathing room
\setlength{\tabcolsep}{8pt} 
% Increase row height slightly to match the image style
\renewcommand{\arraystretch}{0.95} 

\begin{tabular}{llcccc}
\toprule
\textbf{Initial Prompt Type} & \textbf{Prompt Enhancement Method} & \textbf{\begin{tabular}[c]{@{}c@{}}GroundingDINO\\ mIoU\end{tabular}} & \textbf{\begin{tabular}[c]{@{}c@{}}GroundingDINO\\ Conf\end{tabular}} & \textbf{\begin{tabular}[c]{@{}c@{}}YOLO-E\\ mIoU\end{tabular}} & \textbf{\begin{tabular}[c]{@{}c@{}}YOLO-E\\ Conf\end{tabular}} \\
\midrule

\multirow{3}{*}{Underdetailed} 
    & Raw Prompt & 86.60 \% & 83.60 \% & 65.41 \% & 49.23 \% \\
    & Key Object Extraction & 86.20 \% & \textbf{85.00 \%} & \textbf{66.24 \%} & 51.07 \% \\
    & Semantic Category Grounding & \textbf{90.45 \%} & 77.51 \% & 62.21 \% & \textbf{54.81 \%} \\
\midrule

\multirow{3}{*}{Standard} 
    & Raw Prompt & 86.86 \% & 82.85 \% & 63.10 \% & 47.87 \% \\
    & Key Object Extraction & 87.11 \% & \textbf{84.38 \%} & \textbf{64.24 \%} & 49.52 \% \\
    & Semantic Category Grounding & \textbf{90.97 \%} & 77.21 \% & 61.60 \% & \textbf{53.76 \%} \\
\midrule

\multirow{3}{*}{Overdetailed} 
    & Raw Prompt & 59.46 \% & 69.40 \% & \textbf{70.17} \% & \textbf{56.07} \% \\
    & Key Object Extraction & 60.21 \% & \textbf{78.05 \%} & 65.95 \% & 52.49 \% \\
    & Semantic Category Grounding & \textbf{79.98 \%} & 74.43 \% & 62.38 \% & 53.51 \% \\
\midrule

\multirow{3}{*}{Pragmatic Ambiguity} 
    & Raw Prompt & 35.84 \% & 44.59 \% & 9.56 \% & 8.61 \% \\
    & Key Object Extraction & 86.82 \% & \textbf{86.27 \%} & 64.90 \% & 54.57 \% \\
    & Semantic Category Grounding & \textbf{90.99 \% }& 77.51 \% & \textbf{65.66 \%} & \textbf{57.35 \%} \\
\bottomrule
\vspace{-0.9cm}
\end{tabular}
\end{table*}

We first analyze model performance on datasets with simulated prompts, focusing on how different instruction types and prompt post-processing strategies affect detection robustness. Rather than comparing individual object detection models, our analysis emphasizes general trends across prompt formulation and task types in order to identify prompt-related behaviors that influence performance.

\noindent\textbf{Underdetailed.} Underdetailed prompts omit critical visual attributes and resemble short or incomplete user inputs. Under this condition, detection performance shows only a slight improvement across tasks when post-processing methods are applied. These results suggest that when the remaining linguistic cues are already sufficient to identify the target object, prompt refinement provides limited additional benefit under underdetailed user input.

\noindent\textbf{Standard.}
Standard prompts correspond to well-formed, explicit object descriptions that closely resemble how users would naturally refer to objects in everyday interactions. Under this setting, detection performance remains comparable to that observed for underdetailed prompts, suggesting that once a prompt contains sufficient information to uniquely identify the target object, additional descriptive completeness does not substantially improve localization accuracy. This indicates that detection performance saturates when prompts exceed a minimum threshold of semantic clarity. \\
\noindent\textbf{Overdetailed.}
Overdetailed prompts include extensive attribute descriptions specifying multiple visual properties of the target object. Despite appearing more informative, these prompts lead to degraded detection performance relative to standard and underdetailed cases for GD. This suggests that excessive linguistic detail introduces competing or redundant constraints that are difficult for the model to reconcile, particularly in cluttered scenes where multiple objects partially satisfy the described attributes. Prompt post-processing helps mitigate this effect. In particular, Semantic Category Grounding improves GD performance by 20.52\% mIoU compared to raw prompts, while Key Object Extraction increases confidence by 8.65\%. In contrast, YOLO-E achieves its best performance under overdetailed conditions when using raw prompts, with both mIoU and confidence exceeding those obtained with post-processed prompts. \\
\noindent\textbf{Pragmatic Ambiguity.} Pragmatically ambiguous prompts rely on implicit user intent rather than explicit descriptions, making them particularly challenging for language-guided object detection. Under this condition, raw prompts result in severe performance degradation, with GD mIoU dropping to 35.85\% and YOLO-E dropping to 9.56\%, which is substantially lower than performance observed under underdetailed, standard, and overdetailed prompts. Prompt post-processing substantially improves robustness in this setting. For GD, Semantic Category Grounding increases mIoU by 55.15\%, while Key Object Extraction improves confidence by 41.68\%. Similar trends are observed for YOLO-E, where semantic category grounding improves mIoU by 56.10\% and confidence by 48.74\%. These results indicate that explicitly resolving implicit intent is critical for mitigating failures induced by pragmatic ambiguity. 

The results for GD and YOLO-E highlight the dominant role of prompt formulation in determining detection performance. Across most instruction types, variations in localization accuracy are driven primarily by differences in linguistic specification and prompt post-processing rather than changes in visual content. Both models exhibit strong sensitivity to pragmatic ambiguity, while benefiting from prompt refinements that explicitly resolve referential intent. An exception is observed under overdetailed prompts, where YOLO-E achieves its best performance when using raw prompts. This behavior is likely influenced by YOLO-E's recall-oriented detection strategy, which produces a large set of candidate bounding boxes. While higher recall increases the likelihood that at least one prediction satisfies an attribute-rich prompt, it also introduces additional false detections. Together, these observations underscore that prompt formulation interacts closely with model-specific detection behavior, motivating a qualitative analysis of representative failure cases.

\subsection{Failure Case Analysis}
To complement the quantitative results, we present representative failure cases illustrating how prompt-induced errors manifest in practice. This section includes example AR images together with the corresponding prompts and incorrectly predicted bounding boxes, providing a concrete visualization of when open-set object detection models fail and what these failure bounding boxes look like. By examining these cases, we offer qualitative insight into how different prompt formulations lead to incorrect localization, missed detections, or ambiguous predictions in realistic XR scenarios.

\begin{figure}[t]
\centering
\includegraphics[width=\linewidth]{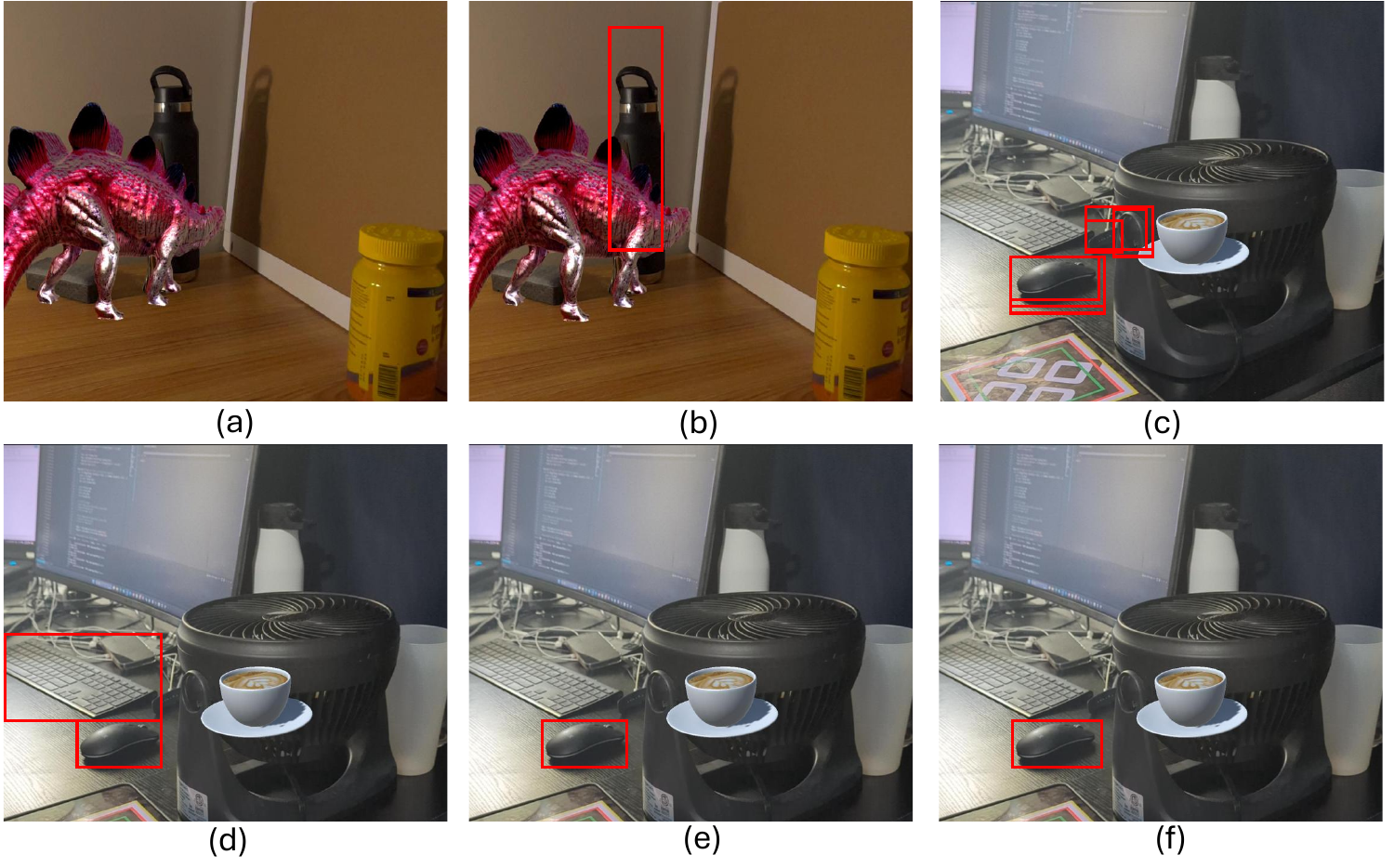}
\vspace{-0.6cm}
\caption{Representative cases of OSOD under different prompt formulations in AR scenes.
% (a) Prompt with pragmatic ambiguity;
% (b) Post-processed ambiguous prompt using semantic category grounding;
% (c) YOLO-E bounding boxes generated from an overdetailed prompt;
% (d) GroundingDINO bounding boxes generated from an overdetailed prompt;
% (e) Post-processed overdetailed prompt evaluated on GroundingDINO;
% (f) Post-processed overdetailed prompt evaluated on YOLO-E.}
(a) Pragmatic ambiguity prompt + no enhancement + GD;
(b) Pragmatic ambiguity prompt + semantic category grounding + GD;
(c) Overdetailed prompt + no enhancement + YOLO-E;
(d) Overdetailed prompt + no enhancement + GD;
(e) Overdetailed prompt + semantic category grounding + YOLO-E;
(f) Overdetailed prompt + semantic category grounding + GD.}
\label{failure}
\vspace{-0.6cm}
\end{figure}

As shown in Figure~\ref{failure}, we present six representative cases illustrating different prompt formulations and OSOD model outputs in AR scenes. Subfigure (a) shows pragmatic ambiguity in GD, where the input prompt “I’m thirsty-pass me the refillable thing I’d drink from with the loop on top” results in no detection. Subfigure (b) applies post-processing using semantic category grounding, yielding the correct object label, “water bottle.” Subfigure (c) illustrates how YOLO-E responds to an overdetailed prompt: “sleek matte-black ergonomic computer mouse with a high rounded hump, two distinct click buttons split by a rubberized ridged scroll wheel, smooth untextured sides, and no visible cable indicating a wireless design.” Due to its recall-oriented strategy, YOLO-E produces a large set of candidate bounding boxes. In contrast, GD in subfigure (d) outputs fewer candidates for the same prompt. Finally, subfigure (e) and (f) show the post-processed versions of the overdetailed prompts for YOLO-E and GD using semantic category grounding, respectively. Both models successfully detect the correct object, “computer mouse,” demonstrating the effectiveness of prompt post-processing.

These examples collectively highlight the sensitivity of OSOD models to prompt formulation and the challenges posed by pragmatic ambiguity, over-specification, and linguistic variability in real-world AR scenarios. The failure cases not only reveal where current OSOD models fall short but also suggest concrete patterns in prompt construction that significantly impact detection performance. Based on these insights, we offer the following prompt design strategies to improve model robustness in XR settings:
\begin{itemize}
    \vspace{-0.2cm}
    \item \textbf{Be explicit and attribute-focused.} Use clear, visual descriptors to reduce ambiguity and guide accurate localization.
    \vspace{-0.2cm}
    \item \textbf{Avoid pragmatic ambiguity.} Indirect expressions such as “I'm thirsty” are often misinterpreted, as current models lack robust intent inference capabilities.
    \vspace{-0.2cm}
    \item \textbf{Limit over-specification.} Excessively detailed prompts can dilute attention or confuse the model. Include only the attributes necessary for disambiguation.
    \vspace{-0.3cm}
\end{itemize}

\section{Limitations and Future Work} 
Despite the insights provided by our analysis, several limitations remain. First, although our dataset includes a substantial number of prompts across various object categories, it does not capture the full range of linguistic expressions that users may employ in open-ended XR environments. In particular, idiomatic, metaphorical, and culturally nuanced language remains underrepresented.

Second, our evaluation is conducted on a relatively small set of images. Broader testing across more diverse environments would offer a more complete assessment of model robustness. Future work will investigate larger-scale benchmarking across a wider range of open-set object detectors, enabling a deeper understanding of natural prompting behavior.

Third, our study does not involve real human users interacting with XR applications. While this controlled setting is sufficient for an initial analysis of prompting strategies and enhancement methods, it does not fully capture how users issue prompts in real XR scenarios, where prompts are influenced by situational context and cognitive constraints. In addition, our study is conducted on single images rather than continuous XR content. In a realistic XR setting, live XR videos could be used to allow the study of temporal and 3D spatial consistency. Future work will evaluate prompt-conditioned OSOD models in real XR applications with human participants to better understand model behavior under realistic user interactions.
\vspace{-0.2cm}
\section{conclusion}

In this work, we investigated prompt-conditioned robustness in OSOD under XR scenarios, moving beyond conventional benchmark evaluations that assume well-formed textual inputs. By simulating user-generated prompts with varying levels of specificity, we show that prompt formulation plays a dominant role in OSOD performance. Both GD and YOLO-E exhibit stable behavior under underdetailed and standard prompts, whereas pragmatically ambiguous prompts cause severe failures across models, and overdetailed prompts introduce competing constraints that degrade performance in GD. Prompt enhancement strategies effectively mitigate ambiguity-induced failures, yielding improvements exceeding 55\% in mIoU and 41\% in average confidence by resolving referential intent. These results highlight the importance of prompt-aware design for language-guided OSOD in XR, suggesting that robust OSOD in practice depends not only on detection architectures but also on how user intent is expressed and enhanced in real-world settings.

\vspace{-0.1cm}
\acknowledgments{
This work was supported in part by NSF grants CSR-2312760, CNS-2112562, and IIS-2231975, NSF CAREER Award IIS-2046072, NSF NAIAD Award 2332744, a Cisco Research Award, a Meta Research Award, Defense Advanced Research Projects Agency Young Faculty Award HR0011-24-1-0001, and the Army Research Laboratory under Cooperative Agreement Number W911NF-23-2-0224. The views and conclusions contained in this document are those of the authors and should not be interpreted as representing the official policies, either expressed or implied, of the Defense Advanced Research Projects Agency, the Army Research Laboratory, or the U.S. Government. This paper has been approved for public release; distribution is unlimited. No official endorsement should be inferred. The U.S.~Government is authorized to reproduce and distribute reprints for Government purposes notwithstanding any copyright notation herein.
}
\vspace{-0.2cm}
\bibliographystyle{abbrv-doi}

\end{document}